\documentclass[conference]{IEEEtran}
\IEEEoverridecommandlockouts    

\usepackage{amsmath}
\usepackage{amssymb}
\usepackage{graphicx}
\usepackage{array}
\usepackage{tabularx}
\usepackage{url}
\usepackage{hyperref}
\graphicspath{{./Figures/}}
\newcolumntype{Y}{>{\raggedright\arraybackslash}X}          
\newcolumntype{L}[1]{>{\raggedright\arraybackslash}m{#1}}   
\newcolumntype{C}[1]{>{\centering\arraybackslash}m{#1}}     

\title{\LARGE \bf Patronus: A Hybrid-Modular Neuro-Symbolic Architecture for Agentic Driving in Long-Tail Scenarios}

\author{
	\parbox{\textwidth}{%
		\centering
		Meghna Pal$^{1}$
	}%
	\thanks{$^{1}$Texas A\&M University, College Station, TX, USA
		{\tt\small meghnapal08@gmail.com, meghnapal08@tamu.edu}}%
}

\begin{document}

	\maketitle
	\thispagestyle{empty}
	\pagestyle{empty}

    
	\begin{abstract}
		Automated vehicles still fail in the long tail of rare and ambiguous situations, where a conservative planner either breaks a rule unsafely or freezes in place. Vision language action models (VLAs) can reason about such scenes, but they are slow, carry no safety guarantee, and lack training data for rare events. We present Patronus, a hybrid neuro-symbolic architecture with an online and an offline part. Online, IRONMAN pairs a deterministic planner and safety layer (TONY) with an asynchronous VLA, Alpamayo 1.5 (JARVIS). JARVIS chooses among trajectories that TONY has already verified, any request outside that set takes effect only if TONY's physics corroborates it, an attentiveness monitor schedules reasoning by scene risk, and an escalation controller resolves deadlocks onboard with short escape sequences executed at creep speed. Offline, Dr.~STRANGE, under development, recreates hard encounters and their counterfactual variations and keeps a reasoning correction for preference fine-tuning only if re-simulation confirms it. IRONMAN runs in closed loop in CARLA with the 10B-parameter VLA live on a single consumer GPU. An audit of 36 episodes (39,207 control ticks) shows where intent and behavior diverge: the veto never triggered, the vehicle executed the default lane-keeping option on 91\% of ticks, and 23\% of ticks were spent waiting for an escape plan. We describe the fixes these findings motivate and an evaluation protocol on Bench2Drive.
	\end{abstract}
    
		\vspace{0.5em}
    
	\section{Introduction}
	\label{sec:introduction}

    Road traffic crashes kill about 1.19 million people every year and are the leading cause of death for children and young people aged 5 to 29, and more than half of those killed are vulnerable road users such as pedestrians, cyclists and motorcyclists \cite{who2023road}. Automated driving is often presented as a way to reduce this toll, but the cases that remain hard are not the common ones. They sit in a long tail of rare and ambiguous situations: a delivery truck double parked in the only open lane, a pedestrian stepping off the curb in the middle of a block, a worker waving traffic through a red light. In these situations a conservative planner tends to fail in one of two ways. It either breaks a rule unsafely, or it refuses to move at all. The second failure, known in crowd navigation as the freezing robot problem \cite{trautman2010unfreezing}, is less visible than a collision but just as real for a vehicle that must share the road.

    Learned planners have moved quickly toward this long tail. End-to-end architectures such as UniAD \cite{hu2023uniad} plan directly from sensor input, and vision language models (VLMs) and vision language action models (VLAs) add a layer of semantic reasoning on top of that \cite{tian2024drivevlm, jiang2024senna, hwang2024emma}. NVIDIA's Alpamayo-R1 pairs trajectory prediction with Chain of Causation reasoning traces aimed specifically at long-tail scenarios \cite{wang2025alpamayor1}, and its open successor, Alpamayo 1.5, is a 10B-parameter VLA built on the Cosmos-Reason2 backbone that accepts natural-language navigation commands and outputs both a 6.4 s trajectory and a text explanation of its reasoning \cite{pavone2026alpamayo15, nvidia2026alpamayo15card}. These models can read a scene the way a human driver does, but three problems keep them from simply taking the wheel.

    The first problem is latency. A large reasoning model cannot run inside a 20 Hz control loop, which is why systems such as DriveVLM-Dual and AsyncDriver run the language model at a low rate beside a fast conventional planner \cite{tian2024drivevlm, chen2024asyncdriver}. In our own implementation, a 4-bit quantized Alpamayo 1.5 on a single consumer GPU takes between 8.4 and 24 s per call (median about 11 s), which is comparable to or longer than the model's own 6.4 s planning horizon. The second problem is assurance. A generative model can produce a confident trajectory that is physically infeasible or unsafe, and established safety architectures such as Simplex \cite{sha2001simplex} and shielding \cite{alshiekh2018shielding} exist precisely because a complex controller should not be the last line of defense. The third problem is data. Long-tail events are by definition rare in driving logs, and a policy trained by imitation drifts into states its training data never covered, where its errors compound \cite{ross2011dagger}. Collecting human corrections for each rare event does not scale.

    This paper presents Patronus, an architecture that addresses these three problems by splitting driving into an online system that is allowed to reason but never allowed to act unchecked, and an offline system that manufactures the rare experience the online system lacks. The online system, IRONMAN (Integrated Reasoning Over Neuro-symbolic Models for Autonomous Navigation), pairs two components. TONY (Time-bounded Optimization for Navigation and Yielding) is a deterministic planner and safety layer that runs every control tick. JARVIS (Joint Action Reasoning and Visual Intent System) wraps Alpamayo 1.5 and runs asynchronously in the background. The central design rule is that JARVIS never drives the vehicle directly. TONY first builds a menu of candidate trajectories and removes every option that fails a kinematic and clearance check, a form of action masking \cite{huang2022actionmasking}. JARVIS then makes a semantic choice from what remains, and any request to leave the menu, such as an emergency acceleration, takes effect only if TONY's own physics corroborates it. Every decision is ranked by a six-tier hierarchy (life, collision, vulnerable road users and emergency vehicles, traffic law, mission, comfort) in which traffic law may be violated only to satisfy one of the three tiers above it. To spend the model's limited reasoning budget where it matters, an attentiveness monitor classifies each surrounding agent into a behavioral state and scores the scene on novelty, reachability, consequence and uncertainty, raising or lowering how often JARVIS is queried. When the vehicle deadlocks, IRONMAN does not wait for a remote human. It escalates to an onboard teleoperator mode in which JARVIS proposes a short, bounded escape sequence that TONY executes at a low escape creep speed (about 3.4 mph).

    The offline system, Dr.~STRANGE (Deep Rollout Simulator for Training Robust Agents in Novel Generative Environments), targets the data problem and is under active development. Its Nightmare Engine recreates the hard situations IRONMAN encounters, together with counterfactual variations of each, and IRONMAN drives them again. A Judge then grades not only what the vehicle did but the reasoning that led to it, locates the step where that reasoning went wrong, and proposes a correction, which is kept only if the re-simulated scenario actually ends safely. These physics-verified corrections become preference pairs for fine-tuning with Direct Preference Optimization (DPO) \cite{rafailov2023dpo}.

    We are explicit about what exists today. IRONMAN is implemented in the CARLA simulator \cite{dosovitskiy2017carla} with range-gated ground-truth perception, and Alpamayo 1.5 runs live in the loop on a single NVIDIA RTX 5080 using about 7.6 GB of GPU memory. Dr.~STRANGE is under active development, and several IRONMAN upgrades, including validity horizons for stale reasoning, a risk-proportional veto and a minimum risk maneuver, are planned. We do not report Bench2Drive scores \cite{jia2024bench2drive}. We instead report an audit of the current implementation, based on a code review and 36 closed-loop episodes (39,207 control ticks). The audit shows that the architecture's intent and its behavior can diverge in ways that matter: the corroborated veto never triggered, because the output parser maps the model's reasoning onto pre-verified menu options before the veto sees it; the vehicle executed the default lane-keeping option on 91\% of control ticks; and the vehicle spent 23\% of all ticks waiting in the escalation request state. We report these findings because they are exactly the failure modes a hybrid reasoning architecture must solve, and because the planned upgrades are designed around them.

    Patronus is a synthesis: its dual-process backbone, verified fallback, language-model scenario generation and preference learning each build on established work, reviewed in Section~\ref{sec:related}. Its intended contributions are therefore not any single one of these ideas, but the following:
    \begin{itemize}
        \item An escalation controller that treats deadlock as a first-class failure and resolves it onboard, by asking the reasoning model for a short, bounded escape sequence that the deterministic layer executes at a low escape creep speed.
        \item A conditional corroboration veto, in which the reasoning model chooses freely among pre-verified trajectories, but any request outside that set takes effect only when the deterministic layer's own physics agrees.
        \item The integration of these mechanisms, together with behavior-driven scheduling of the reasoning model and a six-tier safety hierarchy, into a single online system (IRONMAN) whose encounters are designed to feed a simulator-verified reasoning curriculum (Dr.~STRANGE).
        \item A closed-loop implementation of IRONMAN in CARLA running Alpamayo 1.5 live on a single consumer GPU, and an audit of where that implementation departs from its design, with proposed fixes and an evaluation protocol on Bench2Drive.
    \end{itemize}

    \section{Related Work}
    \label{sec:related}
    
    Patronus draws on five lines of work: language models for driving, dual-process reasoning, safety layers, scenario generation, and learning from feedback. This section reviews each and closes with how Patronus differs (Table~\ref{tab:positioning}).
    
    \subsection{Language Models for Driving}
    \label{sec:rw-vla}
    End-to-end planners such as UniAD \cite{hu2023uniad} map sensor input to trajectories through a single learned network. Vision language models add semantic reasoning: DriveVLM describes and analyzes scenes before planning \cite{tian2024drivevlm}, Senna splits the work so that a VLM makes planning decisions in natural language while an end-to-end model predicts precise trajectories \cite{jiang2024senna}, and EMMA treats driving as a multimodal language modeling task \cite{hwang2024emma}. Alpamayo-R1 pairs trajectory prediction with Chain of Causation reasoning aimed at long-tail scenarios \cite{wang2025alpamayor1}, and Alpamayo 1.5 makes this an open, steerable 10B-parameter VLA \cite{pavone2026alpamayo15, nvidia2026alpamayo15card}. RAD-LAD pairs a real-time language-action planner with a rule-based planner, adding the learned plan as one candidate for a rule-based scorer \cite{ghosh2026radlad}.
    
    \subsection{Dual-Process and Asynchronous Reasoning}
    \label{sec:rw-dual}
    Large models are too slow and too weak at spatial reasoning to control a vehicle on their own. DriveVLM-Dual addresses this by combining the VLM with a traditional driving pipeline \cite{tian2024drivevlm}. AsyncDriver decouples the inference of a language model from a real-time planner, so the expensive model runs at a lower frequency \cite{chen2024asyncdriver}. LeapAD adopts an explicit dual-process design, with an analytic process for reasoning and a heuristic process for fast decisions, and transfers experience from a growing memory bank to the fast process through fine-tuning \cite{mei2024leapad}. These systems invoke reasoning at a fixed rate, continuously, or after a failure.
    
    \subsection{Safety Layers and Rule Hierarchies}
    \label{sec:rw-safety}
    The Simplex architecture runs a complex controller under the supervision of a simple, verified one that takes over when needed \cite{sha2001simplex}, and shielding restricts a learning agent to actions that a formal safety specification permits \cite{alshiekh2018shielding}. Responsibility-Sensitive Safety defines safe distances and proper responses from vehicle kinematics \cite{shalevshwartz2017rss}. Rulebooks order driving rules by priority so that lower rules yield when rules conflict \cite{censi2019rulebooks}. PDM-Closed scores candidate trajectories with a rule-based metric and overrides its choice with emergency braking when a collision is imminent \cite{dauner2023pdm}.
    
    \subsection{Scenario Generation and Simulation}
    \label{sec:rw-scenarios}
    CARLA provides an open simulator for urban driving \cite{dosovitskiy2017carla}, and ASAM OpenSCENARIO standardizes how scenarios are described \cite{asam2024openscenario}. ChatScene uses a language model to generate safety-critical scenarios and turn them into CARLA code \cite{zhang2024chatscene}. NVIDIA OmniDreams, since renamed Cosmos-Dreams \cite{nvidia2026alpamayo2super}, replaces the sensor renderer of a closed-loop simulator with a real-time generative video world model, which generates the next camera frames from the simulator state and the policy's actions and so serves as a reactive environment for evaluating and training driving policies \cite{basant2026omnidreams}. Bench2Drive evaluates closed-loop driving across many interactive scenarios \cite{jia2024bench2drive}.
    
    \subsection{Learning from Feedback}
    \label{sec:rw-feedback}
    Policies trained by imitation drift into states their data never covered; DAgger addresses this by collecting expert labels on the learner's own states \cite{ross2011dagger}. Learning from AI feedback replaces human preference labels with labels from a language model \cite{lee2024rlaif}. DPO learns directly from preference pairs without a separate reward model \cite{rafailov2023dpo}, and LoRA and QLoRA make such fine-tuning affordable on limited hardware \cite{hu2022lora, dettmers2023qlora}. Process supervision rewards each step of a reasoning chain rather than only the final answer \cite{lightman2024prm}.
    
    \subsection{Positioning}
    \label{sec:rw-positioning}
    Table~\ref{tab:positioning} summarizes how Patronus relates to each line of work. None of the reviewed driving systems treats deadlock as an explicit failure mode; the freezing robot problem \cite{trautman2010unfreezing} motivates the escalation controller as a separate contribution.
    
    \begin{table}[htbp]
    \centering
    \footnotesize
    \caption{Established ideas and how Patronus uses them.}
    \label{tab:positioning}
    \begin{tabularx}{\columnwidth}{|C{1.7cm}|Y|Y|}
    \hline
    \textbf{Idea} & \textbf{Prior work} & \textbf{In Patronus} \\ \hline
    Reasoning model & DriveVLM, Senna, EMMA, Alpamayo & Alpamayo 1.5 chooses among pre-verified options; it never drives directly \\ \hline
    Dual process & DriveVLM-Dual, AsyncDriver, LeapAD & Invocation scheduled by scene risk, not a fixed rate \\ \hline
    Safety layer & Simplex, shielding, RSS, rulebooks & Action masking as a shield, a six-tier rulebook and Simplex-style final authority, plus a corroboration veto for requests outside the verified set \\ \hline
    Scenario generation & ChatScene, OmniDreams & Recreates the system's own hard encounters and their counterfactuals in a physics simulator \\ \hline
    Feedback learning & DAgger, RLAIF, DPO, process supervision & Reasoning graded step by step; corrections kept only if re-simulation succeeds \\ \hline
    Deadlock & Freezing robot problem & Onboard escalation with bounded escape sequences \\ \hline
    \end{tabularx}
    \end{table}
    
    
    \section{Framework}
    \label{sec:designandimplementation}

    \begin{figure*}[!htbp]
        \centering
        \includegraphics[width=0.9\linewidth]{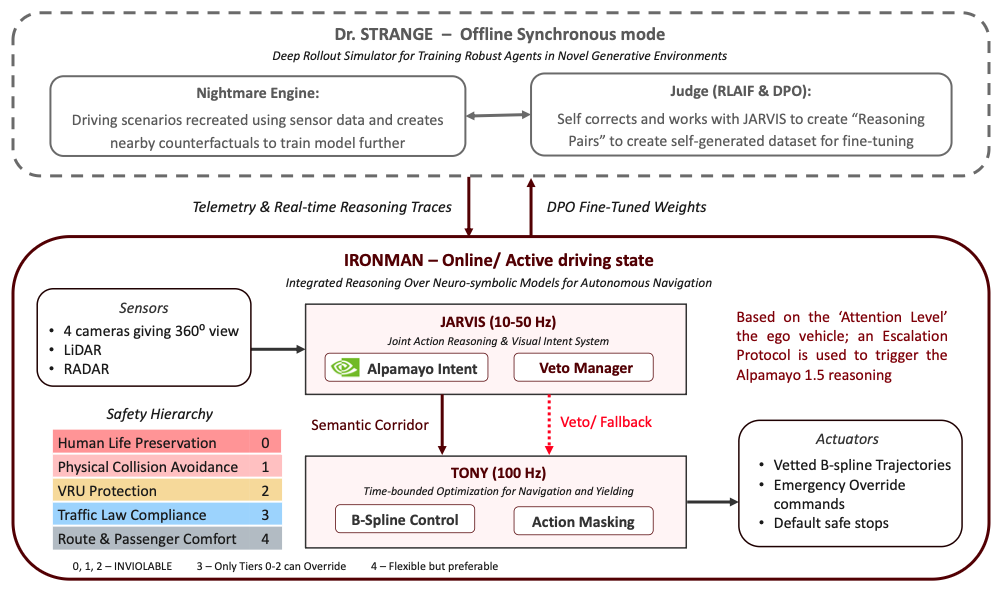}
        \caption{Patronus Framework Overview}
        \label{fig:Framework_diag}
    \end{figure*}
    
    Patronus consists of an online driving system, IRONMAN, and an offline learning system, Dr.~STRANGE. The framework overview is shown in Fig.\ref{fig:Framework_diag}. Table~\ref{tab:status} lists each component and its status. In the parameter lists that follow, each value is marked V (vehicle specification), D (design choice, with the reason given), or H (heuristic: an initial engineering estimate that we expect to replace with physics-derived or learned values, discussed in Section~\ref{sec:limitations}).
    
    \begin{table}[htbp]
    \centering
    \footnotesize
    \caption{Patronus components and implementation status.}
    \label{tab:status}
    \begin{tabularx}{\columnwidth}{|L{2.5cm}|Y|C{1.6cm}|}
    \hline
    \textbf{Component} & \textbf{Role} & \textbf{Status} \\ \hline
    TONY: route planner, trajectory menu, rule engine, controllers & Deterministic planning and safety, every tick & Implemented \\ \hline
    JARVIS (Alpamayo 1.5) & Semantic selection and escape planning & Implemented \\ \hline
    Conditional corroboration veto & Gate on requests outside the menu & Implemented \\ \hline
    Attentiveness monitor & Schedules JARVIS by scene risk & Implemented \\ \hline
    Escalation controller & Onboard deadlock recovery & Implemented \\ \hline
    Preference data pipeline & Records episodes, builds DPO pairs, LoRA training script & Implemented, not yet trained \\ \hline
    Dr.~STRANGE & Offline recreation, counterfactuals and verified reasoning correction & Under development \\ \hline
    IRONMAN upgrades & Minimum risk maneuver, physics thresholds, validity horizons, risk-proportional veto, targeted queries (Section~\ref{sec:futurework}) & Planned \\ \hline
    \end{tabularx}
    \end{table}
    
    \subsection{Control Loop and Scene Representation}
    \label{sec:loop}
    
    IRONMAN runs in CARLA \cite{dosovitskiy2017carla} in synchronous mode with a fixed step of 0.05~s (20~Hz). Each tick executes, in order:
    \begin{enumerate}
        \item Advance the simulator and extract the ground-truth scene state.
        \item Serialize the scene to text and update the ego pose history.
        \item Run the route planner (target speed and upcoming route).
        \item Run the rule engine.
        \item Smooth the route into a base path and generate the trajectory menu.
        \item Update the attentiveness monitor and the escalation controller.
        \item Submit a JARVIS request if one is due; read the latest completed output.
        \item Parse the output into a menu selection and apply the veto.
        \item Apply escalation overrides; compute steering, throttle and brake.
        \item Apply the emergency brake floor, actuate, and record the tick.
    \end{enumerate}
    
    Perception is not modeled. The scene state is read from the simulator and covers agents within 60~m of the ego vehicle (H), with their position, velocity, heading, bounding box and type (vehicle or pedestrian), together with the road context: lane geometry and markings, traffic lights affecting the ego vehicle, and static obstacles. For JARVIS, this state is serialized into scene text containing ego speed and controls, the applicable traffic light, and the ten nearest agents (D: bounds prompt length).
    
    \subsection{TONY: Deterministic Planning and Safety}
    \label{sec:tony}
    
    \subsubsection{Route Planning}
    \label{sec:route}
    A global route to a random destination is computed with CARLA's global route planner. The next 24 route waypoints, prefixed with the ego position, are fitted with an interpolating cubic B-spline and resampled to 30 points to form the base path. Planner parameters:
    \begin{itemize}
        \item \textit{Waypoint spacing}: 2~m (H).
        \item \textit{Minimum destination distance}: 180~m (D: avoids trivially short routes).
        \item \textit{Cruise speed}: 18.6~mph (8.3~m/s) (D: conservative urban speed).
        \item \textit{Junction speed}: 8.9~mph (4.0~m/s) (H).
        \item \textit{Planner creep speed}: 2.2~mph (1.0~m/s) (H).
        \item \textit{Red-light braking onset}: 35~m (H).
        \item \textit{Blocking cone}: 20~m, $\pm 20^\circ$ (H).
        \item \textit{Stuck declaration}: 8~s below 0.7~mph (H).
    \end{itemize}
    
    \subsubsection{Trajectory Menu and Action Masking}
    \label{sec:menu}
    Every tick, TONY generates up to six candidates from the base path (Table~\ref{tab:menu}). Their parameters are listed at the end of this subsection.
    
    \begin{table}[htbp]
    \centering
    \footnotesize
    \caption{Trajectory menu candidates.}
    \label{tab:menu}
    \begin{tabularx}{\columnwidth}{|Y|C{1.7cm}|C{0.9cm}|L{1.5cm}|}
    \hline
    \textbf{Candidate} & \textbf{Lateral offset} & \textbf{Speed} & \textbf{Offered} \\ \hline
    MAINTAIN & 0 & 100\% & Always \\ \hline
    NUDGE\_RIGHT / LEFT & $o_{\text{nudge}}$, Eq.~\eqref{eq:nudge} & 85\% & Always \\ \hline
    LANE\_RIGHT / LEFT & One lane width & 70\% & Outside junctions \\ \hline
    EMERGENCY\_STOP & 0 & 0 & Always, never culled \\ \hline
    \end{tabularx}
    \end{table}
    
    \textit{Offset paths.} Each base-path point is shifted along its local right normal:
    \begin{equation}
    \mathbf{p}'_k = \mathbf{p}_k + o\,\mathbf{n}_k ,
    \label{eq:offset}
    \end{equation}
    where $\mathbf{p}_k$ is the $k$-th base-path point, $\mathbf{n}_k$ is the unit right normal from a central-difference tangent, and $o$ is the signed lateral offset (positive to the right).
    
    \textit{Safety margin.} The margin grows with ego speed:
    \begin{equation}
    m(v) = m_0 + k_m v ,
    \label{eq:margin}
    \end{equation}
    where $v$ is ego speed in m/s, $m_0$ is the base margin, and $k_m$ is the added margin per m/s. At cruise speed, Eq.~\eqref{eq:margin} gives 0.82~m.
    
    \textit{Nudge size.} Each nudge clears the agent that intrudes furthest toward the ego centerline within 20~m ahead:
    \begin{equation}
    o_{\text{nudge}} = \max\!\left(o_{\min},\; \iota + w_e + m(v)\right),
    \label{eq:nudge}
    \end{equation}
    where $\iota$ is that agent's intrusion past the centerline, $w_e$ is the ego half-width, and $o_{\min}$ is the minimum nudge (used alone when nothing intrudes).
    
    \textit{Swept-volume culling.} Each non-stop candidate path is resampled at 1~m into points $\mathbf{s}_k$, with a circle of radius $w_e + m(v)$ at each point. A candidate is culled if, for any agent or static obstacle $j$,
    \begin{equation}
    \min_k \lVert \mathbf{s}_k - \mathbf{x}_j \rVert < w_e + m(v) + w_j ,
    \label{eq:swept}
    \end{equation}
    where $\mathbf{x}_j$ is the obstacle's position and $w_j$ its half-width. Lane changes are also removed if they exceed the available road width.
    
    \textit{Culled candidates stay visible.} They remain in the menu marked BLOCKED, with the blocking agent's identity and distance, so JARVIS can see what was rejected and why. The default is the first feasible non-stop candidate in the order of Table~\ref{tab:menu}, else EMERGENCY\_STOP.
    
    \textit{Relation to prior work.} PDM-Closed generates candidates from centerline offsets and target speeds and selects among them with a rule-based scorer \cite{dauner2023pdm}. IRONMAN delegates the selection among feasible candidates to a reasoning model.
    
    \textit{Menu parameters.}
    \begin{itemize}
        \item Ego half-width $w_e$: 0.93~m (V: CARLA Tesla Model 3).
        \item Base margin $m_0$: 0.40~m (H); margin slope $k_m$: 0.05~s (H).
        \item Minimum nudge $o_{\min}$: 0.8~m (H); nudge look-ahead: 20~m (H).
        \item Sample spacing: 1~m (D: the circle radius is at least 1.33~m, so consecutive circles always overlap and leave no gaps).
        \item Speed factors: 85\% for nudges, 70\% for lane changes (H: lateral maneuvers are flown slower).
    \end{itemize}
    
    \subsubsection{Six-Tier Rule Hierarchy}
    \label{sec:tiers}
    TONY ranks safety requirements in strict priority order, in the spirit of rulebooks \cite{censi2019rulebooks} (Table~\ref{tab:tiers}). A higher tier always overrides a lower one. Traffic law (Tier~3) may be broken only to satisfy Tiers~0 to~2, for example crossing a lane marking to avoid a collision.
    
    \begin{table}[htbp]
    \centering
    \footnotesize
    \caption{Rule hierarchy and current checks.}
    \label{tab:tiers}
    \begin{tabularx}{\columnwidth}{|C{0.5cm}|L{1.5cm}|C{1.3cm}|Y|}
    \hline
    \textbf{Tier} & \textbf{Protects} & \textbf{Violable} & \textbf{Current check} \\ \hline
    0 & Human life & Never & Covered by Tiers 1 and 2 \\ \hline
    1 & Collision avoidance & Never & Clearance $<$ 0.5~m, or forward TTC $<$ 1.5~s $\rightarrow$ emergency brake \\ \hline
    2 & VRUs, emergency vehicles & Never & Pedestrian ahead within 5~m and 2.2~m lateral (junction: 10~m, 5~m) $\rightarrow$ emergency brake \\ \hline
    3 & Traffic law & Only for Tiers 0 to 2 & Warnings: red light; speed $>$10\% above cap \\ \hline
    4 & Mission & Flexible & Route progress \\ \hline
    5 & Comfort & Flexible & Note on hard braking at low speed \\ \hline
    \end{tabularx}
    \end{table}
    
    Threshold reasoning:
    \begin{itemize}
        \item \textit{Clearance 0.5~m, TTC 1.5~s}: H; to be replaced by speed-dependent thresholds (Section~\ref{sec:futurework}).
        \item \textit{TTC corridor}: agents more than 0.5~m behind, or more than 2~m to the side and more than 4~m away, are excluded (D: vehicles overtaking in adjacent lanes are not on a collision course).
        \item \textit{VRU window 5~m, 2.2~m}: D: covers the ego lane ahead; widened at junctions because crossing pedestrians approach from the side.
        \item \textit{Speed cap 31.1~mph (13.9~m/s)}: D: a typical urban limit.
    \end{itemize}
    Context modifiers (school and construction zones, adverse weather, night, nearby emergency vehicles) that tighten speed caps and scale TTC by up to $1.4\times$ are defined but not yet activated. When Tier~1 or~2 fires, a \textit{brake floor} applies full brake after every other component has run, overriding JARVIS, the escalation controller and manual input.
    
    \subsubsection{Low-Level Control}
    \label{sec:control}
    Lateral control uses pure pursuit \cite{coulter1992purepursuit}:
    \begin{align}
    \delta &= \arctan\!\left(\frac{2L\sin\alpha}{L_d}\right), \label{eq:pp}\\
    L_d &= \operatorname{clip}\!\left(k_d v + L_{\min},\; L_{\min},\; L_{\max}\right), \label{eq:lookahead}
    \end{align}
    where $\delta$ is the steering angle, $L$ the wheelbase, $\alpha$ the heading angle to the look-ahead point, and $L_d$ the look-ahead distance. By Eq.~\eqref{eq:lookahead}, $L_d$ grows with speed, which keeps steering smooth at higher speeds. Longitudinal control is a PID controller tracking the executed target speed. Controller parameters:
    \begin{itemize}
        \item Wheelbase $L$: 2.875~m (V: Tesla Model 3).
        \item Look-ahead gain $k_d$: 0.8~s (H).
        \item $L_{\min}$, $L_{\max}$: 3~m, 18~m (H: avoid oscillation at low speed and corner cutting at high speed).
        \item PID gains $K_p$, $K_i$, $K_d$: 0.5, 0.05, 0.1 (H).
    \end{itemize}
    
    \subsection{JARVIS: Asynchronous Semantic Reasoning}
    \label{sec:jarvis}
    
    JARVIS wraps Alpamayo 1.5, a 10B-parameter VLA that outputs a Chain of Causation reasoning trace and a 6.4~s trajectory \cite{pavone2026alpamayo15, nvidia2026alpamayo15card}. Its configuration:
    \begin{itemize}
        \item \textit{Weights}: 4-bit NF4 \cite{dettmers2023qlora} with 16-bit autocast (D: fits beside CARLA on one 16~GB GPU).
        \item \textit{Cameras}: front, left, right and rear, one frame each, $800\times600$, $90^\circ$ FOV, mapped onto the model's four camera slots (D: fewer image tokens, lower latency and memory).
        \item \textit{Ego history}: 8 poses, 0.35~s at 20~Hz (H).
        \item \textit{Prompt}: scene text and menu (plus the context narrative at L2 and above) in the navigation-text slot, at most 1200 characters (D: bounds prompt length and latency).
        \item \textit{Generation}: at most 256 tokens, one trajectory sample (D: bounds latency).
        \item \textit{Sampling}: top-$p$ 0.95, temperature 0.6 (H).
        \item \textit{Execution}: background thread (D: the 20~Hz loop never waits).
    \end{itemize}
    The model's default input is four frames from each of four cameras (front-wide, front-tele, cross-left and cross-right) and 16 ego poses at 10~Hz \cite{nvidia2026alpamayo15card}. This configuration departs from it in frame count, pose history and camera layout: in particular, a rear-facing view occupies one of the slots trained on forward or cross views (Section~\ref{sec:lim-compute}).
    
    \textit{Asynchronous execution.} When a request is due and no inference is running, the main thread snapshots the cameras, ego history and prompt, and starts inference in the background. Until it completes, each tick reuses the latest completed output. Repeated errors trigger a back-off.
    
    \textit{Output parsing.} Only JARVIS's text is used; its predicted trajectory is returned but not executed. The parser checks, in order:
    \begin{enumerate}
        \item EMERGENCY\_BRAKE $\rightarrow$ EMERGENCY\_STOP.
        \item EMERGENCY\_ACCELERATE $\rightarrow$ MAINTAIN at $1.5\times$ its speed.
        \item SWERVE\_LEFT/RIGHT $\rightarrow$ first feasible nudge, then lane change, on that side.
        \item \texttt{TRAJECTORY:} tag naming a feasible candidate $\rightarrow$ that candidate.
        \item Anything else, or a culled choice $\rightarrow$ the default.
    \end{enumerate}
    The executed speed is the minimum of the selected speed and the planner's target speed: JARVIS can slow the vehicle but cannot exceed the planner.
    
    \subsection{Conditional Corroboration Veto}
    \label{sec:veto}
    
    The evidence required depends on what JARVIS asks for (Table~\ref{tab:veto}).
    
    \begin{table}[htbp]
    \centering
    \footnotesize
    \caption{Veto rules. All thresholds are H.}
    \label{tab:veto}
    \begin{tabularx}{\columnwidth}{|L{1.5cm}|Y|Y|}
    \hline
    \textbf{Request} & \textbf{Required evidence} & \textbf{Outcome} \\ \hline
    Emergency brake & None (braking is conservative) & Always approved; logged as unnecessary if nothing within 6~m ahead and no VRU in path \\ \hline
    Emergency accelerate & Agent within 30~m behind, closing faster than 11.2~mph (5.0~m/s) & Else replaced with MAINTAIN \\ \hline
    Swerve & MAINTAIN clearance $<$ 3~m, or VRU within 8~m on the forward path & Else replaced with MAINTAIN \\ \hline
    \end{tabularx}
    \end{table}
    
    Every decision also yields a preference pair, so the veto trains in both directions. When a request is rejected, JARVIS's trace becomes the \textit{rejected} sample and a corrective trace built from ground truth becomes the \textit{chosen} one. When an emergency is corroborated, JARVIS's trace is \textit{chosen} and a synthetic trace that ignores the threat is \textit{rejected}. In the current wiring, the parser (Section~\ref{sec:jarvis}) turns brake and swerve requests into menu entries before the veto sees them, so only EMERGENCY\_ACCELERATE reaches it as an emergency command.
    
    \subsection{Attentiveness Monitor}
    \label{sec:attn}
    
    Running a 10B-parameter model continuously is neither affordable nor useful on an empty street. The attentiveness monitor sets one of four levels, which decides how often JARVIS is consulted (Table~\ref{tab:levels}). A new request starts only after the previous inference has finished.
    
    \begin{table}[htbp]
    \centering
    \footnotesize
    \setlength{\tabcolsep}{3pt}
    \caption{Attentiveness levels: current implementation and design intent.}
    \label{tab:levels}
    \begin{tabularx}{\columnwidth}{|C{0.7cm}|C{1.4cm}|Y|Y|}
    \hline
    \textbf{Level} & \textbf{JARVIS requests} & \textbf{Typical trigger (implemented)} & \textbf{Intended meaning} \\ \hline
    L0 & None & Nothing of note & Routine driving \\ \hline
    L1 & Every 20 ticks (1~s) & Anomalous or uncertain agent; novelty $\geq$ 0.15 & Notice and keep in context; no inference expected \\ \hline
    L2 & Every 5 ticks (0.25~s) & Nearby pedestrian; converging agent; no feasible path; pedestrian in a junction & Inference and action needed; not an emergency \\ \hline
    L3 & Every tick & Converging vehicle with high reachability; active Tier 0 to 2 rule; escalation & Poorly understood situation; continuous active inference \\ \hline
    \end{tabularx}
    \end{table}
    
    \subsubsection{Behavioral States}
    \label{sec:states}
    Each agent is classified every tick from its last 20 snapshots (1~s). States are checked in the order of Table~\ref{tab:states}, most safety-relevant first, and the first match wins.
    
    \begin{table}[htbp]
    \centering
    \footnotesize
    \caption{Behavioral states.}
    \label{tab:states}
    \begin{tabularx}{\columnwidth}{|L{1.9cm}|Y|L{2.1cm}|}
    \hline
    \textbf{State} & \textbf{Criterion} & \textbf{Basis} \\ \hline
    ANOMALOUS & Over the last 4 snapshots (0.15~s): speed change $>$ 6.7~mph (3.0~m/s) or heading change $>35^\circ$; or pedestrian faster than 6.3~mph (2.8~m/s) & Pedestrian: D, about twice walking pace, i.e.\ running. Others: H \\ \hline
    CONVERGING & Constant-velocity projection over 3~s passes within 2.5~m of the ego path & 3~s, 2.5~m: H \\ \hline
    DIVERGING & Distance to ego grew $>$1.5~m over the last 4 snapshots & H \\ \hline
    UNCERTAIN & Pedestrian below 2.7~mph (1.2~m/s), $<$15~m ahead, $<$4~m lateral; or stopped vehicle $<$12~m ahead & D: slower than walking pace suggests pausing or deciding \\ \hline
    PREDICTABLE & None of the above & -- \\ \hline
    \end{tabularx}
    \end{table}
    
    Agents faster than 8.1~mph (3.6~m/s) are never UNCERTAIN, since their intent is clear from motion. The states depend only on kinematics, so a camera-based intent model can replace the estimator through the same interface.
    
    \subsubsection{Agent Properties}
    \label{sec:props}
    Each agent receives four properties.
    
    \textit{Reachability}, how likely the agent is to reach the ego path:
    \begin{equation}
    R_j =
    \begin{cases}
    \min\!\left(1,\; 2\,\mu(s_j)\left(1 - \dfrac{d_j}{d_{\max}}\right)\right), & d_j \le d_{\max},\\[4pt]
    0, & \text{otherwise},
    \end{cases}
    \label{eq:reach}
    \end{equation}
    where $d_j$ is the agent's distance to the ego vehicle, $d_{\max} = 30$~m (D: about 3.6~s of travel at cruise speed), and $\mu(s_j)$ is the reachability multiplier for its state $s_j$ (Table~\ref{tab:mult}). The factor 2 in Eq.~\eqref{eq:reach} lets a converging agent reach $R_j = 1$ within about 13~m, before it is very close.
    
    \textit{Uncertainty}, how many futures are plausible, is a fixed value per state (Table~\ref{tab:mult}). It is computed but not yet used in the level rules.
    
    \begin{table}[htbp]
    \centering
    \footnotesize
    \caption{Per-state multipliers. All values are H: an ordinal ranking of how likely each state is to bring the agent into the ego path.}
    \label{tab:mult}
    \begin{tabularx}{\columnwidth}{|Y|C{1.8cm}|C{1.8cm}|}
    \hline
    \textbf{State} & \textbf{Reach.\ $\mu$} & \textbf{Uncertainty} \\ \hline
    CONVERGING & 0.9 & 0.6 \\ \hline
    ANOMALOUS & 0.7 & 0.9 \\ \hline
    UNCERTAIN & 0.4 & 0.8 \\ \hline
    PREDICTABLE & 0.1 & 0.1 \\ \hline
    DIVERGING & 0.05 & 0.1 \\ \hline
    \end{tabularx}
    \end{table}
    
    \textit{Consequence} $C_j$ is the safety tier (Table~\ref{tab:tiers}) at stake in the worst case: Tier~2 for pedestrians, Tier~1 for converging or anomalous vehicles, and Tier~3 otherwise.
    
    \textit{Novelty}, the fraction of agents whose state changed since the last tick:
    \begin{equation}
    N = \frac{1}{|\mathcal{A}|}\sum_{j \in \mathcal{A}} \mathbf{1}\!\left[s_j(t) \neq s_j(t-1)\right],
    \label{eq:novelty}
    \end{equation}
    where $\mathcal{A}$ is the set of current agents and $\mathbf{1}[\cdot]$ is 1 if the condition holds and 0 otherwise.
    
    \subsubsection{Level Computation}
    \label{sec:levels}
    Each agent contributes a level (Table~\ref{tab:contrib}); the scene level is the maximum contribution, raised by floor rules.
    
    \begin{table}[htbp]
    \centering
    \footnotesize
    \caption{Per-agent level contribution, checked top to bottom. Thresholds are H.}
    \label{tab:contrib}
    \begin{tabularx}{\columnwidth}{|Y|C{1.0cm}|}
    \hline
    \textbf{Condition} & \textbf{Level} \\ \hline
    ANOMALOUS, $R_j > 0.6$, $C_j \le 2$ & L2 \\ \hline
    ANOMALOUS, otherwise & L1 \\ \hline
    Pedestrian with $R_j > 0.3$ & L2 \\ \hline
    CONVERGING, $C_j \le 1$, $R_j > 0.7$ & L3 \\ \hline
    CONVERGING, otherwise & L2 \\ \hline
    UNCERTAIN pedestrian & L2 \\ \hline
    UNCERTAIN, otherwise & L1 \\ \hline
    All others & L0 \\ \hline
    \end{tabularx}
    \end{table}
    
    Floor rules:
    \begin{itemize}
        \item Active Tier 0 to 2 rule, or active escalation $\rightarrow$ L3.
        \item No feasible non-stop candidate $\rightarrow$ at least L2.
        \item Pedestrian within 15~m inside a junction $\rightarrow$ at least L2.
        \item Novelty $N \geq 0.15$ (Eq.~\eqref{eq:novelty}; H) $\rightarrow$ at least L1.
    \end{itemize}
    Two mechanisms support these rules. With \textit{hysteresis}, a level is held for 20 ticks (1~s) before it may drop (D: prevents flicker between levels). With the \textit{context narrative}, a 10~s (H) log of state changes, rule activations and level changes is prepended to the prompt at L2 and above, because Alpamayo keeps no memory between calls.
    
    \textit{Design intent.} The intended level meanings (Table~\ref{tab:levels}, last column) separate noticing (L1) from reasoning (L2) from continuous reasoning (L3) more sharply than the current rules. Novelty is intended to capture \textit{contextual out-of-place-ness}, an agent or object that does not belong where it is (a mattress on a highway, a pedestrian on a freeway), rather than the temporal state change of Eq.~\eqref{eq:novelty}. Aligning the implementation with this intent is planned work (Section~\ref{sec:futurework}).
    
    \subsection{Escalation Controller}
    \label{sec:escalation}
    
    When TONY cannot make progress, IRONMAN escalates to an \textit{onboard teleoperator} mode: JARVIS plans the escape, TONY executes it.
    
    \textit{Trigger.} A stuck timer runs while ego speed is below 1.8~mph (0.8~m/s) and decays three times faster while moving. It resets at red or yellow lights, inside junctions, and with a VRU within 10~m, where waiting is correct. After 10~s, escalation fires if every non-stop candidate has been culled (\textit{all culled}), or if the planner has also been stuck for 10~s (\textit{deep stuck}).
    
    \textit{State machine.}
    \begin{itemize}
        \item \textbf{IDLE}: normal driving, timer running.
        \item \textbf{REQUESTING}: vehicle holds still; JARVIS receives the escape prompt.
        \item \textbf{EXECUTING}: TONY steps through the escape sequence.
        \item \textbf{RETURNING}: control returns to normal driving.
    \end{itemize}
    
    \textit{Escape prompt.} It contains the trigger, lane geometry, the six nearest agents, the full menu including culled options, the safety hierarchy, and a structured response format (situation, causal factors, rule check, options, decision, confidence, justification). It ends with an escape sequence built from these commands:
    \begin{itemize}
        \item \texttt{SWERVE\_LEFT/RIGHT} $d$: ramp the lateral offset to $d$ at escape creep speed.
        \item \texttt{CREEP\_FORWARD} $d$: move $d$ meters at escape creep speed (odometry).
        \item \texttt{STOP}: stop completely.
        \item \texttt{WAIT} $t$: hold for $t$ seconds.
        \item \texttt{DISABLE\_LANE\_CONSTRAINT} / \texttt{ENABLE\_LANE\_CONSTRAINT}: allow or forbid crossing lane markings.
        \item \texttt{HONK}: signal presence (logged).
    \end{itemize}
    
    \textit{Escalation parameters.}
    \begin{itemize}
        \item Stuck speed: 1.8~mph (0.8~m/s) (H).
        \item Escalation timeout: 10~s (H: long enough to rule out ordinary queuing).
        \item Escape creep speed: 3.4~mph (1.5~m/s) (H).
        \item Maximum lateral offset: 4.5~m (D: one 3.5~m lane plus margin; reaches the adjacent lane, no further).
        \item Lateral ramp: 0.05~m per tick, i.e.\ 2.2~mph (1~m/s) (D: no abrupt lateral jumps).
        \item Maximum sequence length: 10 commands (D: keeps plans short and checkable).
    \end{itemize}
    
    The brake floor (Section~\ref{sec:tiers}) stays active throughout, so an escape sequence can never override a Tier~1 or~2 emergency.
    
    \subsection{Preference Data Pipeline}
    \label{sec:pipeline}
    
    Every tick is recorded: scene state, menu, selection, reasoning text, latency, veto decision, escalation state, controls, collisions and rule violations. DPO pairs come from three sources: pairs written by the veto (Section~\ref{sec:veto}), pairs mined from episodes where reasoning preceded a collision or violation (with corrective traces from ground truth), and human feedback records.
    
    Training fine-tunes only the language backbone with LoRA \cite{hu2022lora}, leaving the trajectory decoder frozen. The LoRA update is
    \begin{equation}
    W = W_0 + \frac{\alpha_{\text{LoRA}}}{r} BA ,
    \label{eq:lora}
    \end{equation}
    where $W_0$ is a frozen weight matrix, $A$ and $B$ are trainable low-rank matrices of rank $r$, and $\alpha_{\text{LoRA}}$ scales the update. $B$ starts at zero, so training begins from the unmodified model. The objective is DPO \cite{rafailov2023dpo}:
    \begin{equation}
    \begin{split}
    \mathcal{L} = -\log \sigma\Big(\beta\Big[ &\log\frac{\pi_\theta(y_w \mid x)}{\pi_{\text{ref}}(y_w \mid x)} \\
    &- \log\frac{\pi_\theta(y_l \mid x)}{\pi_{\text{ref}}(y_l \mid x)}\Big]\Big),
    \end{split}
    \label{eq:dpo}
    \end{equation}
    where $x$ is the prompt, $y_w$ and $y_l$ are the chosen and rejected traces, $\pi_\theta$ is the model being trained, $\pi_{\text{ref}}$ is the frozen original, $\sigma$ is the logistic function, and $\beta$ controls how far the model may drift from $\pi_{\text{ref}}$.
    
    \textit{Training configuration} (implemented, not yet run; all values H):
    \begin{itemize}
        \item LoRA rank $r = 16$, scale $\alpha_{\text{LoRA}} = 32$, dropout 0.05.
        \item Target modules: attention and feed-forward projections.
        \item DPO $\beta = 0.1$.
        \item Learning rate $5\times10^{-5}$.
    \end{itemize}
    
    \subsection{Dr.~STRANGE: Learning from Nightmares}
    \label{sec:strange}
    
    On the road, IRONMAN meets each rare event once. Dr.~STRANGE exists to turn every hard encounter into many. It runs offline, apart from the vehicle, so the driving policy never changes while the car is moving: every update is learned, tested and gated before it reaches IRONMAN. Dr.~STRANGE is under active development. Its training end, the preference data pipeline and LoRA training script of Section~\ref{sec:pipeline}, is already in place, and the components below are being built on top of it.
    
    Dr.~STRANGE has one participant and two components. The participant, the \textit{Driver}, is IRONMAN itself (TONY and JARVIS), unchanged: it drives each nightmare exactly as it would on the road and records its reasoning and actions. The \textit{Nightmare Engine} is a scenario generator in CARLA that recreates hard encounters and generates counterfactual variations of them. The \textit{Judge} uses AI feedback, verified in simulation, to grade the Driver's reasoning, propose and verify corrections, and train JARVIS.
    
    \subsubsection{Nightmare Engine}
    \label{sec:nightmare}
    The Nightmare Engine rebuilds the situations IRONMAN found hard and then asks what else could have happened.
    \begin{itemize}
        \item \textit{Source}: hard situations that IRONMAN saves to a transient memory while driving, such as escalations, veto decisions, Tier~1 or~2 activations and high attentiveness levels.
        \item \textit{Recreation}: a language model translates each situation into a structured scenario specification (JSON or ASAM OpenSCENARIO XML \cite{asam2024openscenario}). Actors are anchored to valid road waypoints rather than raw coordinates, which avoids spawn collisions and off-road placement, and fixed simulator and traffic seeds make each replay deterministic.
        \item \textit{Counterfactuals}: each recreated situation is branched into nearby variations, for example a pedestrian stepping out half a second earlier, a lead vehicle braking harder, or an extra actor hidden behind a parked truck. A lesson that holds only for the exact recorded event is not a lesson.
        \item \textit{Curriculum}: variations are ordered from mild to severe, so the Driver faces harder nightmares as it improves.
    \end{itemize}
    Language-model scenario generation in CARLA has been demonstrated before \cite{zhang2024chatscene}. Here it serves a different purpose: every scenario feeds a training loop whose outcomes are checked in simulation. Concurrent work, OmniDreams (since renamed Cosmos-Dreams), takes a different route: a real-time generative video world model renders the camera views of a closed-loop simulator, providing a reactive environment for evaluating and training driving policies, including Alpamayo~1 \cite{basant2026omnidreams}. In the Nightmare Engine, by contrast, both the scene state and the camera images the Driver sees come from a conventional physics simulator, so the Judge's verification step does not depend on views generated by a learned model.
    
    \subsubsection{Judge}
    \label{sec:judge}
    The Judge rests on one principle: \textit{grade the reasoning, not only the action}. In IRONMAN, every action JARVIS takes is the final link in a Chain of Causation. If the action was wrong, some earlier step in that chain was wrong too: a pedestrian went unnoticed, a stopped truck was read as parked, or a menu option was chosen without accounting for a cyclist. Correcting only the action teaches the model what to do in one scene. Correcting the reasoning teaches it why, and that carries over to the counterfactuals. For the same reason, a correct action reached through flawed reasoning is still flagged, because it is likely to fail in the next variation.
    
    The Judge works in five stages:
    \begin{enumerate}
        \item \textbf{Grade.} Review the Driver's reasoning step by step and locate the first step that went wrong. Following process supervision \cite{lightman2024prm}, credit is assigned to individual steps rather than only to the final outcome.
        \item \textbf{Correct.} Rewrite the reasoning from that step onward, arriving at a different action.
        \item \textbf{Verify.} Replay the nightmare with the corrected action. The correction is kept only if the replay ends without collision or rule violation. This builds on learning from AI feedback \cite{lee2024rlaif} with one change: the AI proposes, and the simulator decides.
        \item \textbf{Learn.} Each verified correction and the original reasoning form a preference pair (chosen and rejected), used to fine-tune JARVIS with the objective of Eq.~\eqref{eq:dpo} through LoRA or QLoRA adapters \cite{hu2022lora, dettmers2023qlora}.
        \item \textbf{Gate.} A new adapter replaces the current one only if it does not regress on a held-out suite of nightmares and ordinary driving. Otherwise it is discarded.
    \end{enumerate}
    
    The loop then closes: the improved Driver returns to the Nightmare Engine, which presents harder variations, and the Judge grades again.
    
    \section{Simulations}
    \label{sec:simulations}
    
    This section describes the simulation setup, the compute cost of running a 10B-parameter VLA in the loop, and an audit of the recorded closed-loop episodes. The audit characterizes the current implementation; it is not a benchmark result. The benchmark protocol is given in Section~\ref{sec:protocol}.
    
    \subsection{Experimental Setup and Compute}
    \label{sec:setup}
    
    All recorded episodes used the configuration in Table~\ref{tab:setup}. They ran on a single desktop computer, with CARLA and the quantized Alpamayo 1.5 sharing one consumer GPU.
    
    \begin{table}[htbp]
    \centering
    \footnotesize
    \caption{Simulation and compute setup.}
    \label{tab:setup}
    \begin{tabularx}{\columnwidth}{|L{2.0cm}|Y|}
    \hline
    \textbf{Item} & \textbf{Setting} \\ \hline
    \multicolumn{2}{|l|}{\textit{Simulation}} \\ \hline
    Simulator & CARLA 0.9.15 \cite{dosovitskiy2017carla}, synchronous mode, 0.05~s step (20~Hz) \\ \hline
    Map, weather & Town01, clear noon \\ \hline
    Traffic & 30 vehicles and 15 pedestrians, CARLA Traffic Manager \\ \hline
    Ego vehicle & Tesla Model 3 \\ \hline
    Perception & Ground truth, agents within 60~m \\ \hline
    Routes & Random destination at least 180~m away \\ \hline
    Episode end & Session end, or 4,000 ticks (200~s simulated) \\ \hline
    \multicolumn{2}{|l|}{\textit{Compute}} \\ \hline
    GPU & NVIDIA RTX 5080, 16~GB \\ \hline
    Framework & PyTorch 2.7 with CUDA 12.8 \\ \hline
    Alpamayo 1.5 & 4-bit NF4 weights, about 7.6~GB of GPU memory in use \\ \hline
    CARLA & Same GPU, server-side rendering disabled \\ \hline
    \end{tabularx}
    \end{table}
    
    \textit{Real-time factor.} Because the simulator is synchronous, it runs only as fast as the control loop allows. We measure this with the real-time factor
    \begin{equation}
    \rho = \frac{T_{\text{sim}}}{T_{\text{wall}}},
    \label{eq:rtf}
    \end{equation}
    where $T_{\text{sim}}$ is the simulated time and $T_{\text{wall}}$ is the wall-clock time over the same episodes. Across the recorded episodes, $\rho \approx 0.49$: the simulation ran at about half real time.
    
    \textit{Inference latency.} In wall-clock time, JARVIS inference took between 8.4 and 24.0~s, with a median and mean of 10.9~s. At the real-time factor of Eq.~\eqref{eq:rtf}, the median corresponds to roughly 5~s of simulated time, still comparable to the model's 6.4~s planning horizon.
    
    \subsection{Implementation Audit}
    \label{sec:audit}
    
    \subsubsection{Recorded Data}
    \label{sec:data}
    Table~\ref{tab:episodes} summarizes the recorded episodes.
    
    \begin{table}[htbp]
    \centering
    \footnotesize
    \caption{Recorded episodes.}
    \label{tab:episodes}
    \begin{tabularx}{\columnwidth}{|Y|C{2.0cm}|}
    \hline
    \textbf{Quantity} & \textbf{Value} \\ \hline
    Episodes & 36 \\ \hline
    Control ticks & 39,207 \\ \hline
    Simulated driving time & 32.7~min \\ \hline
    Mean ego speed & 6.9~mph \\ \hline
    Ticks stationary ($<$0.2~mph) & 48.9\% \\ \hline
    Episodes with a collision & 4 \\ \hline
    Tier~1 activations (ticks) & 4 (11) \\ \hline
    Tier~2 activations (ticks) & 1 (213) \\ \hline
    \end{tabularx}
    \end{table}
    
    Usable reasoning was not available on every tick, mainly because the vehicle is held still until the first inference completes and because of intermittent inference errors on the shared GPU.
    
    \subsubsection{Reasoning Content}
    \label{sec:content}
    JARVIS outputs are single-sentence driving rationales in Alpamayo's native Chain of Causation style. Representative outputs (verbatim):
    \begin{itemize}
        \item \textit{Routine driving}:
        \begin{itemize}
            \item ``Keep lane since the lane is clear ahead''
            \item ``Keep distance to the lead vehicle since it is directly ahead in our lane''
            \item ``Adapt speed for the right curve since the lane bends right ahead''
            \item ``Stop for the stop sign ahead''
        \end{itemize}
        \item \textit{Semantic judgment that TONY cannot make on its own}:
        \begin{itemize}
            \item Yielding: ``Slow down to yield to the pedestrian crossing ahead since they are entering our lane''
            \item Stopping: ``Stop for the stationary lead vehicle blocking our lane ahead''
            \item Merging: ``Maintain lane and wait for a gap to merge left because a car is coming up and then passes in our left lane, blocking our merge path; once it clears, we can merge left as planned for our leftward route''
            \item Nudging: ``Nudge to the right to clear the green road obstacle blocking the center of our lane''
        \end{itemize}
    \end{itemize}
    
    \subsubsection{Executed Trajectories and Veto}
    \label{sec:executed}
    Table~\ref{tab:executed} shows the executed menu candidate per tick. MAINTAIN dominated, with nudges and emergency stops making up most of the remainder. The corroborated veto (Section~\ref{sec:veto}) received no emergency command in any tick, consistent with the parser bypass described there.
    
    \begin{table}[htbp]
    \centering
    \footnotesize
    \caption{Executed menu candidate per tick.}
    \label{tab:executed}
    \begin{tabularx}{\columnwidth}{|Y|C{1.4cm}|C{1.2cm}|}
    \hline
    \textbf{Candidate} & \textbf{Ticks} & \textbf{Share} \\ \hline
    MAINTAIN & 35,607 & 90.8\% \\ \hline
    EMERGENCY\_STOP & 2,537 & 6.5\% \\ \hline
    NUDGE\_RIGHT & 534 & 1.4\% \\ \hline
    NUDGE\_LEFT & 524 & 1.3\% \\ \hline
    LANE\_LEFT & 5 & $<$0.1\% \\ \hline
    \end{tabularx}
    \end{table}
    
    \subsubsection{Escalation}
    \label{sec:escresults}
    The escalation controller (Section~\ref{sec:escalation}) fired 8 times, and the vehicle spent 22.9\% of all ticks in the REQUESTING state. Three escalations reached EXECUTING, with TONY carrying out a JARVIS escape sequence; the other five did not receive a usable escape sequence and remained in REQUESTING. The causes are analyzed in Section~\ref{sec:limitations}.
    
    \subsubsection{Summary}
    \label{sec:auditsummary}
    \begin{itemize}
        \item \textit{Deterministic layer}: worked as designed. It culled infeasible candidates, fell back safely, and triggered Tier~1 and~2 emergency braking.
        \item \textit{Reasoning layer}: when available, it often understood the scene correctly, and in three of eight deadlocks produced an escape sequence that TONY executed.
        \item \textit{Interface between them}: remains the weak point. The veto never received a request, and most escalations stalled while waiting for a usable response.
    \end{itemize}
    Strengthening this interface between reasoning and executed action is the central goal of the planned upgrades (Section~\ref{sec:futurework}) and Dr.~STRANGE (Section~\ref{sec:strange}).
    
    \subsection{Evaluation Protocol}
    \label{sec:protocol}
    
    Quantitative evaluation will use Bench2Drive \cite{jia2024bench2drive}, which runs on CARLA and tests closed-loop driving on 220 short routes of about 150~m each, five for each of 44 interactive scenarios. Table~\ref{tab:b2d} lists the metrics we will report.
    
    \begin{table}[htbp]
    \centering
    \footnotesize
    \caption{Bench2Drive metrics \cite{jia2024bench2drive}.}
    \label{tab:b2d}
    \begin{tabularx}{\columnwidth}{|L{2.0cm}|Y|}
    \hline
    \textbf{Metric} & \textbf{Definition} \\ \hline
    Driving Score & Route completion multiplied by infraction penalties, averaged over routes \\ \hline
    Success Rate & Share of routes completed without any infraction \\ \hline
    Efficiency & Ego speed relative to surrounding traffic \\ \hline
    Comfortness & Whether motion stays within expert smoothness bounds \\ \hline
    Multi-ability & Success on merging, overtaking, giving way, traffic signs and emergency braking \\ \hline
    \end{tabularx}
    \end{table}
    
    Planned ablations, each run on the full route set:
    \begin{itemize}
        \item \textit{TONY only}: no JARVIS, to measure what reasoning adds.
        \item \textit{Fixed-interval JARVIS}: attentiveness monitor disabled, to measure the value of risk-based scheduling.
        \item \textit{No veto} and \textit{no escalation}: each mechanism removed in turn.
    \end{itemize}
    Besides the benchmark metrics, we will report inference count and latency, the share of ticks with usable reasoning, and the share of JARVIS outputs that reach the controls. Because IRONMAN uses ground-truth perception, its scores are comparable to other planners with privileged inputs, not to sensor-based end-to-end systems.
    
    \section{Limitations}
    \label{sec:limitations}
    
    This section states what the current system assumes, what constrains it, and what does not yet work. Issues are reported for the implementation described in Section~\ref{sec:designandimplementation}; directions for addressing them are given in Section~\ref{sec:futurework}.
    
    \subsection{Perfect Perception}
    \label{sec:lim-perception}
    
    IRONMAN isolates planning and reasoning by reading the scene directly from the simulator, which removes an entire class of real-world failure. Positions, velocities, headings and sizes are exact, with no false positives, missed detections or perception delays. Agents are filtered by range (60~m) only, not by visibility, so TONY knows about a pedestrian hidden behind a parked truck, which a real vehicle would not. Agent types are also coarse: agents are labeled only as vehicle or pedestrian, so bicycles and motorcycles count as vehicles, and the Tier~2 VRU rules and the VRU consequence score do not apply to cyclists. Finally, JARVIS reasons over rendered camera images while TONY uses ground truth, so the two layers never disagree because of a perception error, a case a deployed system would have to handle.
    
    As a result, scores obtained with ground-truth perception are an upper bound on what the planning and reasoning stack can achieve with real sensors, and thresholds chosen without noise (Section~\ref{sec:designandimplementation}) may be too tight once perception is imperfect. The behavioral-state interface (Section~\ref{sec:states}) is designed so that a camera-based intent model can replace the kinematic estimator, and injecting noise and occlusion into the ground truth is a planned step toward real perception.
    
    \subsection{Hardware and Software Constraints}
    \label{sec:lim-compute}
    
    Running a 10B-parameter VLA beside a simulator on one consumer GPU forced several compromises, each with a cost to performance.
    
    \textit{GPU memory.} The 16~GB GPU is shared with CARLA, so the model runs with 4-bit NF4 weights and CARLA's server rendering is disabled. The lower numerical precision may affect Alpamayo's reasoning quality, which has not been measured.
    
    \textit{Latency.} Inference takes 8 to 24~s. It runs in a background thread, the latest output is reused every tick, and attentiveness scheduling limits how often the model is called, but the vehicle still acts on reasoning about a scene several seconds old, with no check of how stale that reasoning is.
    
    \textit{Model inputs.} To bound latency, each call receives one frame per camera, 8 ego poses and at most 1200 prompt characters, and generates at most 256 tokens and one trajectory sample. The model therefore gets no motion cues from video, its input departs from its training configuration, and long prompts are truncated. In addition, the four $90^\circ$ CARLA cameras (front, left, right and rear) are mapped onto slots that the model expects to hold front-wide, front-tele, cross-left and cross-right views \cite{nvidia2026alpamayo15card}. The rear view has no counterpart in that rig, and the other viewpoints differ from it.
    
    \textit{Software stack.} The new GPU architecture requires PyTorch 2.7 with CUDA 12.8 and pinned dependency versions. Intermittent CUDA runtime errors still occur during inference; a back-off handles them, but reasoning is unavailable while they persist.
    
    \textit{Synchronous simulation.} The fixed 20~Hz step makes stepping deterministic, but the simulation runs at about half real time ($\rho \approx 0.49$, Eq.~\eqref{eq:rtf}), so experiments are slow and wall-clock latency does not map directly to simulated time.
    
    \subsection{Known Issues}
    \label{sec:lim-issues}
    
    A code review of the implementation, together with the audit of Section~\ref{sec:audit}, identified the following issues, grouped by component.
    
    \subsubsection{TONY}
    \label{sec:lim-tony}
    \textit{Path generation.} The road-width check compares a full lane-width offset against half a lane plus a margin, so lane changes are always removed and the menu is limited to MAINTAIN, nudges and stop. Offsets also apply at full size from the first path point, with no curvature or steering check, which creates abrupt lateral demand and does not guarantee that the offset path is feasible. Swept-volume culling uses each obstacle's half-width at its current position, so long vehicles are under-covered and moving agents are not projected forward.
    
    \textit{Safety rules and control.} The 1.5~s TTC threshold and the clearance distances are fixed, so safety margins do not scale with speed. Context modifiers are defined but never set, and Tier~0 has no check of its own, so school zones, weather and night do not change behavior. The PID derivative acts on the error with no jerk limit, which causes control spikes when the target speed changes and reduces comfort.
    
    \textit{Rerouting.} The planner reroutes and resets its stuck timer at 8~s, so the 10~s deep-stuck escalation trigger can never fire; a blockage leads to a new random destination instead.
    
    \subsubsection{JARVIS Interface}
    \label{sec:lim-jarvis}
    \textit{Prompt.} The menu and context are placed in the navigation-text slot, which the model was trained to read as short navigation commands, so it answers in its native style and often ignores the requested menu format. Prompts are truncated to 1200 characters from the end, so the menu, which comes last, can be cut off entirely. The context narrative makes this worse: it adds one line per flagged agent per tick, so the 10~s window floods and pushes the scene and menu past the truncation point.
    
    \textit{Output.} The parser matches exact keywords, so correct reasoning phrased in other words (``Nudge to the right'') is not recognized. The model's predicted trajectory, its main output, is discarded. Because the executed speed is capped at the planner's target, EMERGENCY\_ACCELERATE has no effect on speed.
    
    \subsubsection{Veto and Attentiveness}
    \label{sec:lim-veto}
    \begin{itemize}
        \item \textit{Parser bypass.} Brake and swerve requests become menu entries before reaching the veto, so only acceleration is ever checked, and the veto did not trigger in the recorded episodes.
        \item \textit{Uncertainty unused.} It is computed for every agent but plays no part in the level rules.
        \item \textit{Novelty.} It measures how many agents changed state, not contextual out-of-place-ness as intended (Section~\ref{sec:levels}).
        \item \textit{Anomaly thresholds.} A 6.7~mph (3.0~m/s) speed change within 0.15~s implies about 20~m/s$^2$ ($\approx$2~g), so only extreme events register as anomalous.
        \item \textit{Level semantics.} L1 still triggers inference every second, although its intended meaning is to notice without reasoning.
    \end{itemize}
    
    \subsubsection{Escalation}
    \label{sec:lim-escalation}
    Four issues together explain why five of the eight escalations in Section~\ref{sec:escresults} stalled in REQUESTING. First, the escape prompt is consumed on the tick it is issued, so if no inference starts on that tick, the prompt is lost. Second, the escape prompt is far longer than 1200 characters, so the command list and the required \texttt{ESCAPE\_SEQUENCE} format, which come last, are cut off. Third, the response handler accepts the latest cached output, which may be a normal-driving answer from before the escape prompt; if it cannot be parsed, the escalation is abandoned. Fourth, there is no timeout, so if only error or placeholder outputs arrive, the vehicle waits in REQUESTING indefinitely.
    
    A separate issue affects escalations that do execute: swerve targets are absolute positions rather than relative moves, so \texttt{SWERVE\_LEFT 3} followed by \texttt{SWERVE\_RIGHT 3} ends 3~m right of the route instead of back at its center.
    
    \subsubsection{Data and Training}
    \label{sec:lim-data}
    \begin{itemize}
        \item \textit{Text-only preferences.} DPO pairs contain reasoning text only; images and the trajectory decoder are not part of the preference signal.
        \item \textit{Prompt mismatch.} The system prompt used to build training pairs is not the prompt the model sees at inference.
        \item \textit{Cumulative collision flag.} Once a collision occurs, every later tick in the episode is marked as colliding, which inflates collision counts when mining episodes.
        \item \textit{No fixed seeds.} Simulator and traffic randomness are not seeded, so episodes cannot be replayed exactly. Deterministic replay is a prerequisite for the Nightmare Engine (Section~\ref{sec:nightmare}).
        \item \textit{Untrained pipeline.} Preference fine-tuning has not been run, so there is no evidence yet that it improves driving.
    \end{itemize}
    
    \subsection{Scope of Evaluation}
    \label{sec:lim-scope}
    
    The evaluation is narrow. All episodes use one town (Town01), clear weather, CARLA's non-adversarial background traffic and no fixed route set, and together they cover only about 33 minutes of simulated driving. Bench2Drive results (Section~\ref{sec:protocol}) are pending, so no claim is made about driving performance relative to other methods. Finally, no real-world data or vehicles have been used, and the gap between CARLA and real roads is not addressed.
    
    \section{Future Work}
    \label{sec:futurework}
    
    The planned work follows directly from the limitations in Section~\ref{sec:limitations} and falls into four groups: making TONY's safety layer physics-based, getting JARVIS's reasoning to the controls on time and under verification, aligning the attentiveness monitor with its design intent, and moving toward realistic perception, benchmark evaluation and learning. Each group below first states the limitations it addresses and then describes the planned work.
    
    \subsection{Physics-Based Safety Layer}
    \label{sec:fw-safety}
    
    \textit{Limitations addressed.} TONY stops by default when no candidate is acceptable, its TTC and distance thresholds are fixed rather than scaled with speed, and its anomaly thresholds register only extreme events of about 2~g. A minimum risk maneuver replaces the default stop, and physics-derived thresholds replace the fixed values.
    
    \subsubsection{Minimum Risk Maneuver}
    \label{sec:fw-mrm}
    SAE J3016 defines a minimal risk condition as a stable, stopped state that reduces crash risk when a trip cannot continue \cite{sae2021j3016}. TONY's current fallback, a stop, matches that definition but is not always the least harmful choice. When no candidate is acceptable, TONY will instead pick the least harmful option. Candidates are compared first by the highest tier they would violate (Table~\ref{tab:tiers}), then by the deceleration each requires:
    \begin{equation}
    a_{\text{req}} = \frac{v^2}{2\,d_{\text{gap}}},
    \label{eq:areq}
    \end{equation}
    where $v$ is ego speed and $d_{\text{gap}}$ is the gap to the nearest obstacle on that candidate's path. A candidate is physically possible only if $a_{\text{req}}$ is within the vehicle's braking capability. Hysteresis prevents the choice from oscillating between ticks.
    
    \subsubsection{Physics-Derived Thresholds}
    \label{sec:fw-thresholds}
    The heuristic thresholds marked H in Section~\ref{sec:designandimplementation} will be replaced by speed-dependent ones, following the reasoning of Responsibility-Sensitive Safety \cite{shalevshwartz2017rss}:
    \begin{align}
    d_{\text{stop}} &= v\,t_{\text{sys}} + \frac{v^2}{2a_{\text{brake}}} + d_0, \label{eq:dstop}\\
    T_h &= t_{\text{sys}} + \frac{v}{a_{\text{brake}}}, \label{eq:headway}
    \end{align}
    where $d_{\text{stop}}$ is the distance needed to stop, $t_{\text{sys}}$ is the system reaction latency, $a_{\text{brake}}$ is the braking deceleration, $d_0$ is the standstill margin, and $T_h$ is the minimum safe time headway. The values of $t_{\text{sys}}$, $a_{\text{brake}}$ and $d_0$ will be calibrated for the simulated vehicle.
    
    Anomaly detection will use each agent's own history instead of fixed global thresholds:
    \begin{equation}
    z_j = \frac{|u_j - \bar{u}_j|}{\operatorname{sd}(u_j)},
    \label{eq:zscore}
    \end{equation}
    where $u_j$ is the agent's current speed or heading, and $\bar{u}_j$ and $\operatorname{sd}(u_j)$ are its mean and standard deviation over its recent history. ``Sudden'' is then judged relative to how that agent normally moves.
    
    \subsection{Timely and Verified Reasoning}
    \label{sec:fw-reasoning}
    
    \textit{Limitations addressed.} The vehicle acts on reasoning that may be several seconds old, every query pays the full inference latency, the veto is keyed to command names and is bypassed by the parser, and correct reasoning often fails to reach the controls because the menu format is ignored, prompts are truncated and escape prompts are dropped. Validity horizons handle stale reasoning, targeted queries reduce latency, a risk-proportional veto replaces the name-based one, and a rebuilt reasoning-to-action interface fixes the rest.
    
    \subsubsection{Validity Horizons}
    \label{sec:fw-validity}
    A reasoning result describes the scene at the moment its snapshot was taken. The earliest time agent $j$ could close its distance to the ego vehicle is
    \begin{equation}
    t_j = \frac{-v_j + \sqrt{v_j^2 + 2a_j d_j}}{a_j},
    \label{eq:ti}
    \end{equation}
    where $d_j$ is its distance, $v_j$ its closing speed, and $a_j$ its worst-case acceleration (as $a_j \to 0$, $t_j \to d_j / v_j$). The result stays valid for
    \begin{equation}
    T_{\text{valid}} = \min_j t_j - \tau,
    \label{eq:tvalid}
    \end{equation}
    where $\tau$ is the inference latency. If $T_{\text{valid}} \le 0$ when the result arrives, the scene may have changed beyond what the reasoning considered, and the result is discarded rather than executed.
    
    \subsubsection{Targeted Queries}
    \label{sec:fw-queries}
    Many decisions are binary, such as whether it is safe to pass a stopped vehicle now. Instead of requesting a full reasoning trace, TONY will ask JARVIS a single go or no-go question with a short generation budget, which reduces latency where it matters most.
    
    \subsubsection{Risk-Proportional Veto}
    \label{sec:fw-veto}
    The veto will scale the evidence it requires with the risk of an action rather than its name, and it will check every executed action rather than only named emergency commands. Low-risk actions, slowing down and stopping, will be approved directly, since they are conservative. Medium-risk actions, nudges within the lane, will need only menu feasibility. High-risk actions, which include accelerating, changing lanes, crossing markings and escape sequences, will need physical corroboration from TONY.
    
    \subsubsection{Reasoning-to-Action Interface}
    \label{sec:fw-interface}
    The audit showed that correct reasoning does not reliably reach the controls. The planned changes are:
    \begin{itemize}
        \item \textit{Prompt structure.} Place the menu and response format first and the context narrative last, and record each agent once per state change rather than once per tick, so that truncation removes history rather than options.
        \item \textit{Tolerant parsing.} Map free-form reasoning onto menu candidates by meaning (for example, ``nudge to the right'' to NUDGE\_RIGHT), not only by exact keywords.
        \item \textit{Richer actions.} Add Alpamayo's own predicted trajectory to the menu as one more candidate, subject to the same culling, as RAD-LAD does for a rule-based scorer \cite{ghosh2026radlad}. Also give JARVIS a library of basic actions with situation-based guidance (how hard to brake, how far to swerve), each checked by TONY.
        \item \textit{Escalation repairs.} Hold the escape prompt until an inference actually starts, reject responses computed before the prompt, add a timeout to REQUESTING, make swerve offsets relative, and let the planner's stuck state trigger escalation instead of rerouting.
        \item \textit{Speed authority.} Allow a corroborated EMERGENCY\_ACCELERATE to exceed the planner's target speed.
    \end{itemize}
    
    \subsection{Attentiveness Aligned with Design Intent}
    \label{sec:fw-attn}
    
    \textit{Limitations addressed.} L1 still triggers inference, novelty measures state changes rather than out-of-place-ness, the uncertainty score is unused, and anomaly thresholds are fixed.
    
    The levels will follow their intended meanings (Table~\ref{tab:levels}, last column): L1 will record the event in context without invoking JARVIS, L2 will request inference, and L3 will keep inference running while the threat lasts. Novelty will measure whether an agent or object belongs where it is, such as a mattress on a highway, using JARVIS's scene understanding rather than state changes. The per-agent uncertainty score will enter the level rules, and the fixed anomaly thresholds will be replaced by the per-agent score of Eq.~\eqref{eq:zscore}.
    
    \subsection{Perception, Evaluation and Learning}
    \label{sec:fw-learning}
    
    \textit{Limitations addressed.} Perception is perfect, the camera rig does not match the model's training rig, the evaluation is narrow with no benchmark results, episodes cannot be replayed exactly, and the preference pipeline has not been trained.
    
    \subsubsection{Perception and Evaluation}
    \label{sec:fw-eval}
    Noise, missed detections and visibility-based occlusion will be injected into the ground truth, and a cyclist class will be added so that the Tier~2 rules cover cyclists. The kinematic behavioral-state estimator will then be replaced by a camera-based intent model through the same interface. Because Alpamayo 1.5 accepts a flexible number of cameras \cite{pavone2026alpamayo15}, the rear camera will be dropped and the remaining views matched to the slots the model was trained on. Simulator and traffic seeds will be fixed for reproducibility, and the Bench2Drive protocol and ablations of Section~\ref{sec:protocol} will be run across its towns and weather conditions.
    
    \subsubsection{Completing Dr.~STRANGE}
    \label{sec:fw-strange}
    Development of Dr.~STRANGE (Section~\ref{sec:strange}) proceeds in order: deterministic replay of recorded encounters, counterfactual generation in the Nightmare Engine, the Judge's grade, correct and verify loop, and finally DPO fine-tuning behind the regression gate. Two data issues will be fixed first: matching the training prompt to the inference prompt, and replacing the cumulative collision flag with per-tick collision events.
    
    \section{Conclusion}
    \label{sec:conclusion}
    
    This paper presented Patronus, a hybrid architecture in which a 10B-parameter VLA reasons about the scene while a deterministic planner keeps final authority over every action. Its intended contributions are an escalation controller that resolves deadlock onboard, a conditional corroboration veto, and the integration of these, together with risk-based scheduling and a six-tier safety hierarchy, into an online system (IRONMAN) whose encounters feed a simulator-verified reasoning curriculum (Dr.~STRANGE).
    
    IRONMAN runs today in closed loop in CARLA, with the VLA reasoning live on a single consumer GPU. Running it showed both what the architecture makes possible and where intent and behavior diverge. The reasoning layer often read scenes correctly, yielding to pedestrians, stopping for blocked lanes, and in three of eight deadlocks producing an escape sequence that the deterministic layer carried out. At the same time, the reasoning model answers on a timescale of seconds while the vehicle acts in fractions of one, the output parser can route around the veto it was meant to feed, and the escalation controller can stall while waiting for a usable answer.
    
    These are not incidental bugs. They are the central problems that any hybrid of slow reasoning and fast control must solve, and validity horizons, a risk-proportional veto, physics-derived thresholds and a minimum risk maneuver are designed around them (Section~\ref{sec:futurework}). The next steps are to implement these fixes, complete Dr.~STRANGE so that IRONMAN learns from its own hardest encounters, and report closed-loop results on Bench2Drive, so that each intended contribution can be tested rather than argued.
    
    \bibliographystyle{ieeetr}
    \bibliography{root}
    
\end{document}